%% file: main.tex
\documentclass[runningheads]{llncs}

\usepackage{eccv}

\usepackage{eccvabbrv}

\usepackage{wrapfig}
\usepackage{enumitem}
\usepackage{multirow}
\usepackage[misc]{ifsym}
\usepackage{graphicx}
\usepackage{booktabs}

\usepackage[accsupp]{axessibility}  

\usepackage{hyperref}

\usepackage{orcidlink}

\let\titleold\title
\renewcommand{\title}[1]{\titleold{#1}\newcommand{\thetitle}{#1}}
\def\maketitlesupplementary
   {
    {\newpage
        \centering
        \Large
        \textbf{\thetitle}\\
        \vspace{0.5em}Supplementary Material \\
        \vspace{1.0em}
        }
   }

\begin{document}

\title{MotionLCM: Real-time Controllable Motion Generation via Latent Consistency Model}

\titlerunning{MotionLCM}

\author{%
  Wenxun Dai$^{1,2}$\orcidlink{0009-0004-1465-3739}, Ling-Hao Chen$^{2}$\thanks{Project lead. ~~~\Letter~Corresponding author.}\orcidlink{0000-0002-2528-6178}, Jingbo Wang$^{3\text{\Letter}}$\orcidlink{0009-0005-0740-8548}, Jinpeng Liu$^{1,2}$\orcidlink{0009-0009-8689-8248} \\ Bo Dai$^{3\text{\Letter}}$\orcidlink{0000-0003-0777-9232}, Yansong Tang$^{1,2}$\orcidlink{0000-0002-1534-4549}
}

\authorrunning{W.~Dai et al.}

\institute{
    $^{1}$Shenzhen Key Laboratory of Ubiquitous Data Enabling, Tsinghua Shenzhen International Graduate School \ $^{2}$Tsinghua University \ $^{3}$Shanghai AI Laboratory \\
    \email{\{wxdai2001, thu.lhchen, wangjingbo1219, liu.jinpeng.55\}@gmail.com} \\
    \email{\{doubledaibo, tangyansong15\}@gmail.com} \\
    Project page: \url{https://dai-wenxun.github.io/MotionLCM-page}
}

\maketitle

\begin{figure}[h]
\begin{center}
\includegraphics[width=0.92\linewidth]{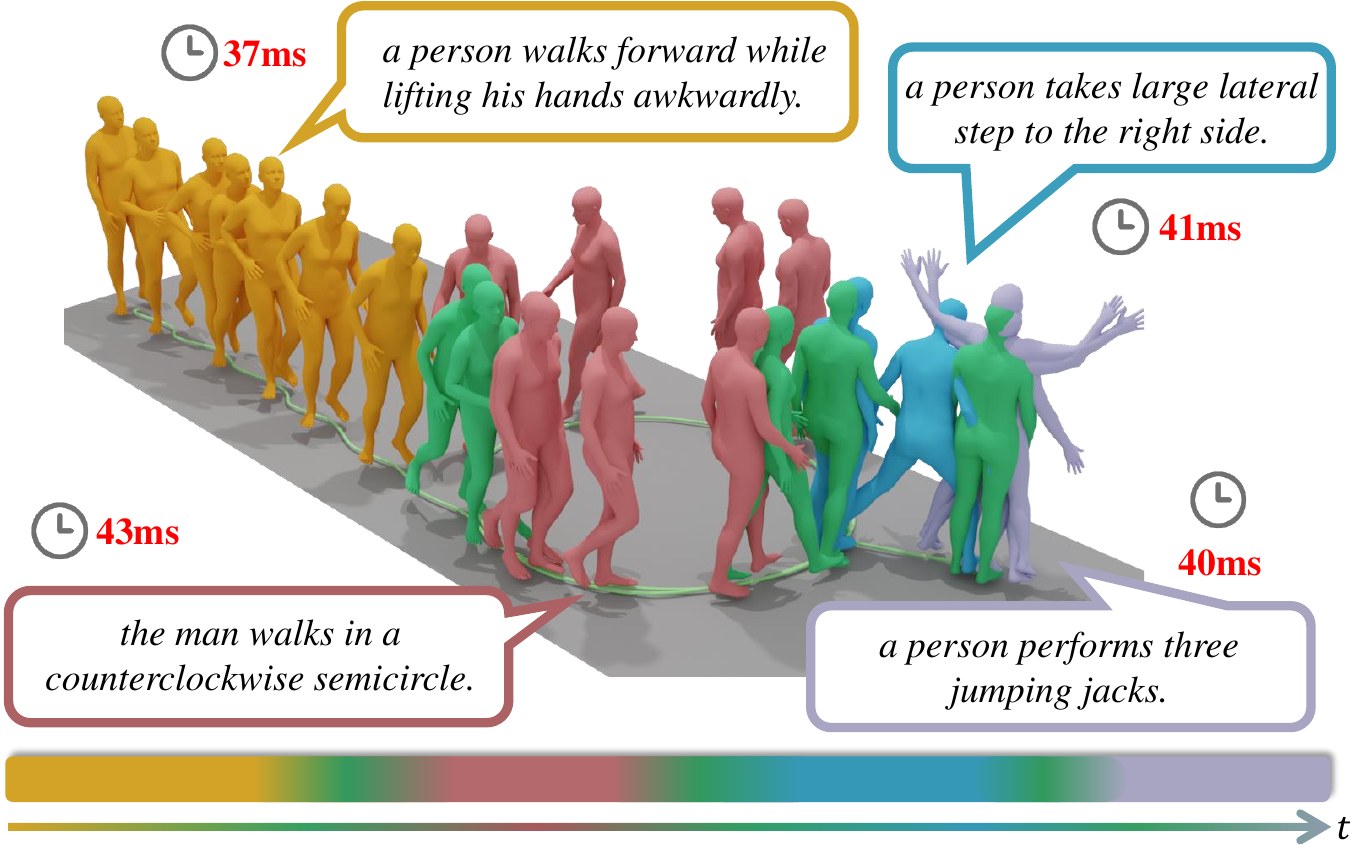}
\end{center}
\caption{We propose MotionLCM, a real-time controllable motion latent consistency model. Our model uses the last few frames of the previous motion as temporal control signals to autoregressively generate the next motion in real-time under different text prompts. Green blocks denote the junctions. The numbers in red are the inference time.}
\label{figures/teaser}
\end{figure}

\input{sections/0_abstract}
\input{sections/1_introduction}
\input{sections/2_related_work}
\input{sections/3_method}
\input{sections/4_experiments}

\input{sections/5_conclusion}

\bibliographystyle{splncs04}
\bibliography{main}

\input{sections/appendix}

\end{document}

%% file: sections/0_abstract.tex
\begin{abstract}

This work introduces MotionLCM, extending controllable motion generation to a real-time level. Existing methods for spatial-temporal control in text-conditioned motion generation suffer from significant runtime inefficiency. To address this issue, we first propose the motion latent consistency model (MotionLCM) for motion generation, building on the motion latent diffusion model. By adopting one-step (or few-step) inference, we further improve the runtime efficiency of the motion latent diffusion model for motion generation. To ensure effective controllability, we incorporate a motion ControlNet within the latent space of MotionLCM and enable explicit control signals (\textit{i.e.}, initial motions) in the vanilla motion space to further provide supervision for the training process. By employing these techniques, our approach can generate human motions with text and control signals in real-time. Experimental results demonstrate the remarkable generation and controlling capabilities of MotionLCM while maintaining real-time runtime efficiency.

\keywords{Text-to-Motion \and Real-time Control \and Consistency Model}

\end{abstract}

%% file: sections/1_introduction.tex
\section{Introduction}

\begin{wrapfigure}{r}{0.52\textwidth}
    \centering
    \includegraphics[width=\linewidth]{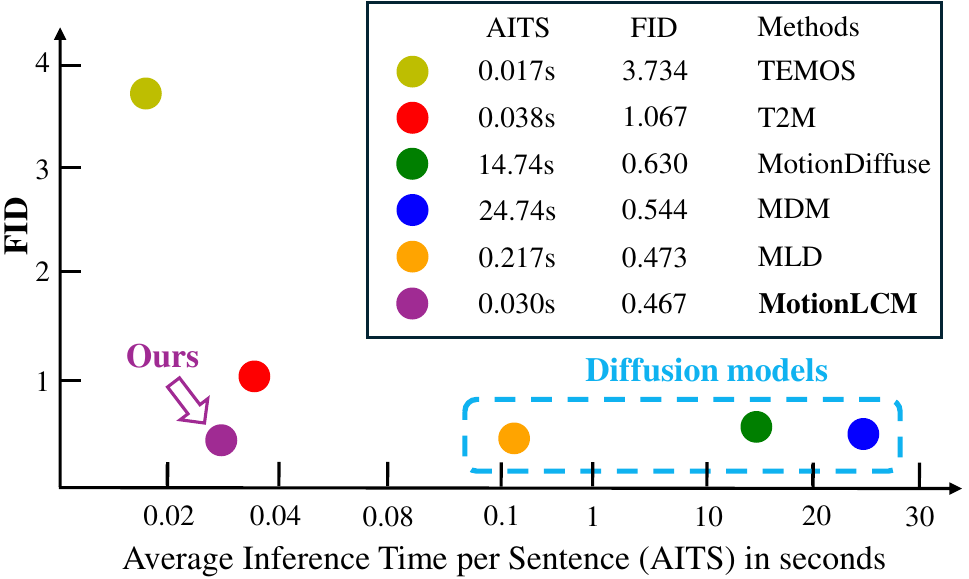}
    \caption{Comparison of the inference time costs on HumanML3D~\cite{humanml3d}. We compare the AITS and FID metrics with five SOTA methods. The closer the model is to the origin the better. Diffusion-based models are indicated by the blue dashed box. Our MotionLCM achieves real-time inference speed while ensuring high-quality motion generation.}
\end{wrapfigure}

Text-to-motion generation~(T2M) has attracted increasing attention~\cite{text2action, temos, mdm, humantomato, momask} due to its important roles in many applications~\cite{humanise, unihsi}. Previous attempts mainly focus on GANs~\cite{text2action, dvgans}, VAEs~\cite{action2motion, actor, temos, teach} and diffusion models~\cite{motiondiffuse, mdm, mld, mofusion, physdiff, diffprior, promotion} via pairwise text-motion data~\cite{ntu, kit, uestc, amass, babel, humanml3d, flag3d, interx} and achieve impressive generation results. Existing approaches~\cite{motiondiffuse, mdm, mld} mainly take diffusion models~\cite{ddpm1, ddpm2, ddpm3, sd} as a base generative model, owing to their powerful ability to model motion distribution. However, these diffusion fashions inevitably require considerable sampling steps for motion synthesis during inference, even with some sampling acceleration methods~\cite{ddim}. Specifically, MDM~\cite{mdm} and MLD~\cite{mld} require $\sim$24s and $\sim$0.2s to generate a high-quality motion sequence. Such low efficiency blocks the applications of generating high-quality motions in various real-time scenarios.

In addition to the language description itself serving as a coarse control signal, another line of research focuses on controlling the motion generation with spatial-temporal constraints~\cite{diffprior, gmd, omnicontrol}. Although these attempts enjoy impressive controlling ability in the T2M task, there still exists a significant gap towards real-time applications. For example, OmniControl~\cite{omnicontrol} exhibits a relatively long inference time, $\sim$81s per sequence. Therefore, trading-off between generation quality and efficiency is a challenging problem. As a result, in this paper, we target the real-time controllable motion generation research problem.

Recently, the concept of consistency models~\cite{cm, lcm} has been introduced in image generation, resulting in significant progress by enabling efficient and high-fidelity image synthesis with a minimal number of sampling steps (\textit{e.g.}, 4 steps \textit{vs.} 50 steps). These properties perfectly align with our goal of accelerating motion generation without compromising generation quality. Therefore, we propose MotionLCM (Motion \underline{L}atent \underline{C}onsistency \underline{M}odel) distilled from the motion latent diffusion model, MLD~\cite{mld}, to tackle the low-efficiency problem in diffusion sampling. To the best of our knowledge, we introduce consistency distillation into the motion generation area \textit{for the first time} and accelerate motion generation to a real-time level via latent consistency distillation~\cite{lcm}.

Here, in MotionLCM, we are facing another challenge on how to control motions with spatial-temporal signals (\textit{i.e.}, initial motions) in the latent space. Previous methods~\cite{diffprior, omnicontrol} model human motions in the vanilla motion space and can manipulate the motion directly in the denoising process. However, for our latent-diffusion-based MotionLCM, it is non-trivial to feed the control signals into the latent space. This is because the latent has no explicit motion semantics, which cannot be manipulated directly by the control signals. Inspired by the notable success of~\cite{controlnet} in controllable image generation~\cite{sd}, we introduce a motion ControlNet to control motion generation in the latent space. However, the na\"ive motion ControlNet is not sufficient to provide supervision for the control signals. The main reason is the lack of explicit supervision in the motion space. Therefore, during the training phase, we decode the predicted latent through the frozen VAE~\cite{vae} decoder into the vanilla motion space to provide explicit control supervision on the generated motion. Thanks to the powerful one-step inference capability of MotionLCM, the latent generated by MotionLCM can significantly facilitate control supervision both in the latent space and motion space for training the motion ControlNet compared to MLD~\cite{mld}.

With our key designs, our proposed MotionLCM successfully enjoys the balance between generation quality and efficiency in controllable motion generation. Before delivering into detail, we sum up our core contributions as follows.

\begin{itemize}
\item We propose the Motion Latent Consistency Model (MotionLCM) via consistency distillation on the motion latent diffusion model extending controllable motion generation to a real-time level. 
\item Building upon our achievement of real-time motion generation, we introduce a motion ControlNet, enabling high-quality controllable motion generation.
\item Extensive experimental results show that MotionLCM enjoys a good balance of generation quality, controlling capability, and real-time efficiency.
\end{itemize} 

%% file: sections/2_related_work.tex
\section{Related Work}

Generating human motions can be divided into three main fashions according to inputs: motion synthesis (1) without any condition~\cite{csgn, bayesian, modi, mdm}, (2) with some given multi-modal conditions, such as action labels~\cite{action2motion, actor, inr, actformer, multiact, case}, textual description~\cite{text2action, dvgans, seq2seq, jl2p, t2g, hier, humanml3d, motionclip, temos, tm2t, motiondiffuse, teach, mdm, humanise, mld, mofusion, physdiff, t2mgpt, diffprior, gmd, tmr, motiongpt, unihsi, humantomato, tlcontrol, momask, promotion, emdm, stmc, flowmdm, move, stablemofusion, freemotion, smoodi, laserhuman, motionllm}, audio or music~\cite{fact, danceformer, bailando, edge, finedance, lodge}, (3) with user-defined trajectories~\cite{diffprior, gmd, omnicontrol, intercontrol, tlcontrol, holden2016deep, pfnn, amdm, aamdm}. To generate diverse, natural, and high-quality human motions, many generative models have been explored by~\cite{jl2p, dvgans, temos, rodinhd, gaussiancube}. Recently, diffusion-based models significantly improved the motion generation performance and diversity~\cite{motiondiffuse, mdm, mld, mofusion, humanmac, regennet, guo2024generative} with stable training. Specifically, MotionDiffuse~\cite{motiondiffuse} represents the first text-based motion diffusion model that provides fine-grained instructions on body parts and achieves arbitrary-length motion synthesis with time-varied text prompts. MDM~\cite{mdm} introduces a motion diffusion model that operates on raw motion data, enabling both high-quality generation and generic conditioning that together comprise a good baseline for new motion generation tasks. Based on MDM~\cite{mdm}, OmniControl~\cite{omnicontrol} integrates flexible spatial-temporal control signals across different joints by combining analytic spatial guidance and realism guidance into the diffusion model, ensuring that the generated motion closely conforms to the input control signals. The work most relevant to ours is MLD~\cite{mld}, which introduces a motion latent-based diffusion model to enhance generation quality and reduce computational resource requirements. The key idea is training a VAE~\cite{vae} for motion embedding, followed by implementing latent diffusion~\cite{sd} within the learned latent space. However, these diffusion fashions inevitably require considerable sampling steps for motion synthesis during inference, even with some sampling acceleration methods~\cite{ddim}. Thus, we propose MotionLCM, which not only guarantees high-quality controllable motion generation but also achieves real-time runtime efficiency.

%% file: sections/3_method.tex
\section{Method}

In this section, we first briefly introduce preliminaries about latent consistency models in~\cref{subsection: Preliminaries}. Then, we describe how to conduct latent consistency distillation for motion generation in~\cref{subsection: MotionLCM: Motion Latent Consistency Model}, followed by our implementation of motion control in latent space in~\cref{subsection: Controllable Motion Generation in Latent Space}. The overall pipeline is illustrated in~\cref{figures/pipeline_v2}.

\subsection{Preliminaries}
\label{subsection: Preliminaries}

\begin{figure}[t]
    \centering
    \includegraphics[width=.9\columnwidth]{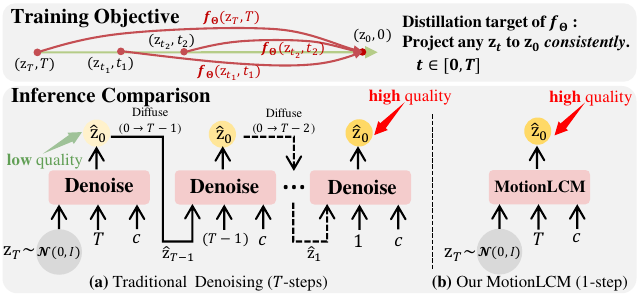}
    \caption{The training objective of consistency distillation is to learn a consistency function $\textbf{{\textit{f}}}_{\mathbf{\Theta}}$, initialized with the parameters of a pre-trained diffusion model (\textit{e.g.}, MLD~\cite{mld}). This function $\textbf{{\textit{f}}}_{\mathbf{\Theta}}$ should projects any points (\textit{i.e.}, $\mathbf{z}_t$) on the ODE trajectory to its solution (\textit{i.e.}, $\mathbf{z}_0$). Once the pre-trained model~\cite{mld} is distilled, unlike the traditional denoising model~\cite{motiondiffuse, mdm} that requires considerable sampling steps, our MotionLCM can generate high-quality motion sequences with one-step sampling and further improve the generation quality through multi-step inference.}
\end{figure}

The Consistency Model (CM)~\cite{cm} introduces a kind of efficient generative model designed for efficient one-step or few-step generation. Given a Probability Flow ODE~(\textit{a.k.a.} PF-ODE) that smoothly converts data to noise, the CM is to learn the function $\textbf{\textit{f}}(\cdot, \cdot)$ that maps any points on the ODE trajectory to its origin distribution (\textit{i.e.}, the solution of the PF-ODE). The consistency function is formally defined as $\textbf{\textit{f}}: (\mathbf{x}_t, t) \longmapsto \mathbf{x}_\epsilon$, where $t \in [\epsilon, T]$, $T > 0$ is a fixed constant and $\epsilon$ is a small positive number to avoid numerical instability. According to~\cite{cm}, the consistency function should satisfy the \textit{self-consistency property}:
\begin{equation}
    \textbf{{\textit{f}}}(\mathbf{x}_t, t) = \textbf{{\textit{f}}}(\mathbf{x}_{t^{\prime}}, t^{\prime}), \forall t, t^{\prime} \in [\epsilon, T]. 
    \label{eq:self-consistency}
\end{equation}
As shown in~\cref{eq:self-consistency}, the self-consistency property indicates that for arbitrary pairs of $(\mathbf{x}_{t}, t)$ on the same PF-ODE trajectory, the outputs of the model should be consistent. The goal of a parameterized consistency model $\textbf{\textit{f}}_{\mathbf{\Theta}}$ is to learn a consistency function from data by enforcing the self-consistency property in~\cref{eq:self-consistency}. To ensure that $\textbf{\textit{f}}_{\mathbf{\Theta}}(\mathbf{x}, \epsilon) = \mathbf{x}$, the consistency model $\textbf{\textit{f}}_{\mathbf{\Theta}}$ is parameterized as,
\begin{equation}
    \textbf{{\textit{f}}}_{\mathbf{\Theta}}(\mathbf{x}, t) = c_{\text{skip}}(t)\mathbf{x} + c_{\text{out}}(t)\textbf{{\textit{F}}}_{\mathbf{\Theta}}(\mathbf{x}, t),
    \label{eq:skip}
\end{equation}
where $c_{\text{skip}}(t)$ and $c_{\text{out}}(t)$ are differentiable functions with $c_{\text{skip}}(\epsilon) = 1$ and $c_{\text{out}}(\epsilon) = 0$, and $\textbf{{\textit{F}}}_{\mathbf{\Theta}}(\cdot, \cdot)$ is a deep neural network to learn the self-consistency. The CM trained from distilling the knowledge of pre-trained diffusion models is called \textit{Consistency Distillation}. The consistency loss is defined as follows,
\begin{equation}
    {\cal L}({\bf \Theta}, {\bf \Theta}^{-}; {\bf \Phi}) = \mathbb{E} \left[d\left(\textbf{{\textit{f}}}_{{\bf \Theta}}(\mathbf{x}_{t_{n+1}}, t_{n+1}), \textbf{{\textit{f}}}_{{\bf \Theta}^{-}}(\hat{\mathbf{x}}^{{\bf \Phi}}_{t_n}, t_n)\right) \right],
    \label{eq:consistency_loss}
\end{equation}
where $d(\cdot, \cdot)$ is a chosen metric function for measuring the distance between two samples. $\textbf{{\textit{f}}}_{\bf \Theta}(\cdot, \cdot)$ and $\textbf{{\textit{f}}}_{{\bf \Theta}^{-}}(\cdot, \cdot)$ are referred to as ``online network'' and ``target network'' according to~\cite{cm}. Besides, ${\bf \Theta}^{-}$ is updated with the exponential moving average (EMA) of the parameters of ${\bf \Theta}$~\footnote{EMA operation: ${\bf \Theta}^{-} \leftarrow \texttt{sg}(\mu{\bf \Theta}^{-} + (1 - \mu) {\bf \Theta})$, where $\texttt{sg}(\cdot)$ denotes the stopgrad operation and $\mu$ satisfies $0 \leq \mu < 1$.}. In~\cref{eq:consistency_loss}, $\hat{\mathbf{x}}^{{\bf \Phi}}_{t_n}$ is a one-step estimation of $\mathbf{x}_{t_{n}}$ from $\mathbf{x}_{t_{n+1}}$, which is formulated as,
\begin{equation}
    \hat{\mathbf{x}}^{{\bf \Phi}}_{t_n} \leftarrow \mathbf{x}_{t_{n+1}} + (t_n - t_{n+1}){\mathbf{\Phi}}(\mathbf{x}_{t_{n+1}}, t_{n+1}, \emptyset),
\end{equation}
where $\bf \Phi$ is a one-step ODE solver applied to PF-ODE.

The Latent Consistency Model (LCM)~\cite{lcm} learns the self-consistency property in the latent space ${D}_\mathbf{z}=\{(\mathbf{z}, \mathbf{c})|\mathbf{z}={\cal E}(\mathbf{x}), (\mathbf{x}, \mathbf{c}) \in {D}\}$, where $D$ denotes the dataset, ${\bf c}$ is the given condition, and ${\cal E}$ is the pre-trained encoder. Compared to CMs~\cite{cm} using the numerical continuous PF-ODE solver~\cite{karras}, LCMs~\cite{lcm} adopt the discrete-time schedule~\cite{ddim, dpm1, dpm2} to adapt to Stable Diffusion~\cite{sd}. Instead of ensuring consistency between adjacent time steps $t_{n+1} \rightarrow t_{n}$, LCMs~\cite{lcm} are designed to ensure consistency between the current time step and $k$-step away, \ie, $t_{n+k} \rightarrow t_{n}$, thereby significantly reducing convergence time costs. As classifier-free guidance (CFG)~\cite{cfg} plays a crucial role in synthesizing high-quality text-aligned images, LCMs integrate CFG into the distillation as follows,
\begin{equation}
    \hat{\mathbf{z}}^{{{\bf \Phi}, w}}_{t_n} \leftarrow \mathbf{z}_{t_{n+k}} + (1+w) {\mathbf{\Phi}}(\mathbf{z}_{t_{n+k}}, t_{n+k}, t_{n},\mathbf{c}) - w{\mathbf{\Phi}}(\mathbf{z}_{t_{n+k}}, t_{n+k}, t_{n}, \emptyset),
\end{equation}
where $w$ denotes the CFG scale which is uniformly sampled from $[w_{\text{min}}, w_{\text{max}}]$ and $k$ is the skipping interval. To efficiently perform the above $k$-step guided distillation, LCMs augment the consistency function to $\textbf{\textit{f}}: (\mathbf{z}_t, t, w, \mathbf{c}) \longmapsto \mathbf{z}_0$, which is also the form adopted by our MotionLCM.

\subsection{MotionLCM: Motion Latent Consistency Model}
\label{subsection: MotionLCM: Motion Latent Consistency Model}

\begin{figure}[t]
    \centering
    \includegraphics[width=1.0\columnwidth]{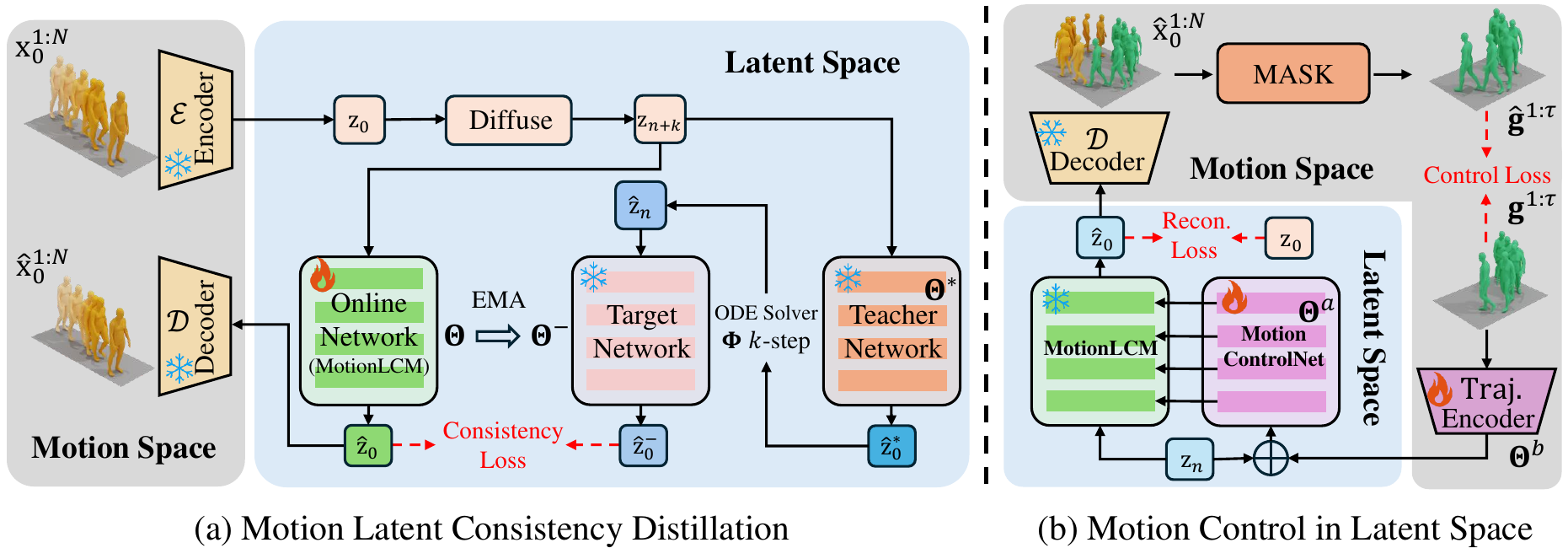}
    \caption{\textbf{The overview of MotionLCM.} (a) \textit{Motion Latent Consistency Distillation (\cref{subsection: MotionLCM: Motion Latent Consistency Model}).} Given a raw motion sequence $\mathbf{x}^{1:N}_0$, a pre-trained VAE~\cite{vae} encoder first compresses it into the latent space, then a forward diffusion operation is performed to add $n+k$ steps of noise. Then, the noisy $\mathbf{z}_{n+k}$ is fed into the online network and teacher network to predict the clean latent. The target network takes the $k$-step estimation results of the teacher output to predict the clean latent. To learn self-consistency, a loss is applied to enforce the output of the online network and target network to be consistent. (b) \textit{Motion Control in Latent Space (\cref{subsection: Controllable Motion Generation in Latent Space}).} With the powerful MotionLCM trained in the first stage, we incorporate a motion ControlNet into the MotionLCM to achieve controllable motion generation. Furthermore, we leverage the decoded motion to explicitly supervise the spatial-temporal control signals (\textit{i.e.}, initial poses $\mathbf{g}^{1:\tau}$).}
    \label{figures/pipeline_v2}
\end{figure}

\textbf{Motion compression into the latent space.} Motivated by~\cite{cm, lcm}, we propose MotionLCM (Motion \underline{L}atent \underline{C}onsistency \underline{M}odel) to tackle the low-efficiency problem in motion diffusion models~\cite{mdm, motiondiffuse}, unleashing the potential of LCM in the motion generation task. Similar to MLD~\cite{mld}, our MotionLCM adopts a consistency model in the motion latent space. We choose MLD~\cite{mld} as the underlying diffusion model to distill from. We aim to achieve few-step (2$\sim$4) and even one-step inference without compromising motion quality. In MLD, an autoencoder ($\cal E$, $\cal D$) is first trained to compress a high dimensional motion into low dimensional latent vectors $\mathbf{z}={\cal E}(\mathbf{x})$, which are then decoded to reconstruct the motion as $\hat{\mathbf{x}}={\cal D}(\mathbf{z})$. Training diffusion models in the motion latent space greatly reduces the computational resources compared to the vanilla diffusion models trained on raw motion sequences (\textit{i.e.}, motion space) and speeds up the inference process. Thus, we effectively leverage the motion latent space for consistency distillation.

\noindent \textbf{Motion latent consistency distillation.} An overview of our motion latent consistency distillation is described in \cref{figures/pipeline_v2}(a). A raw motion sequence $\mathbf{x}^{1:N}_0=\{\mathbf{x}^i\}_{i=1}^{N}$ is a sequence of human poses, where $N$ is the number of frames. We follow~\cite{humanml3d} to use the redundant motion representation for our experiments, which is widely used in previous work~\cite{mdm, motiondiffuse, mld}. As shown in~\cref{figures/pipeline_v2}(a), given a raw motion sequence $\mathbf{x}^{1:N}_0$, a pre-trained VAE~\cite{vae} encoder first compresses it into the latent space, $\mathbf{z}_0={\cal E}(\mathbf{x}_0)$. Then, a forward diffusion operation with $n+k$ steps is conducted to add noise on $\mathbf{z}_0$, where $k$ is the skipping interval illustrated in~\cref{subsection: Preliminaries}. The noisy $\mathbf{z}_{n+k}$ is fed to the frozen teacher network and trainable online network to predict the clean $\hat{\mathbf{z}}^{*}_{0}$, and $\hat{\mathbf{z}}_{0}$. The target network uses the cleaner $\hat{\mathbf{z}}_{n}$ obtained by a $k$-step ODE solver $\bf \Phi$, such as DDIM~\cite{ddim} to predict the $\hat{\mathbf{z}}^{-}_{0}$. Since the classifier-free guidance (CFG)~\cite{cfg} is essential for condition alignment in diffusion models~\cite{sd, mdm, mld}, we integrate CFG into the distillation,
\begin{equation}
    \hat{\mathbf{z}}_{n} \leftarrow \mathbf{z}_{n+k} + (1+w) {\mathbf{\Phi}}(\mathbf{z}_{n+k}, t_{n+k}, t_{n},\mathbf{c}) - w{\mathbf{\Phi}}(\mathbf{z}_{n+k}, t_{n+k}, t_{n}, \emptyset),
\end{equation}
where $\mathbf{c}$ is the text condition and $w$ denotes the guidance scale. To ensure the self-consistency property defined in \cref{eq:self-consistency}, the latent consistency distillation loss ${\cal L}_{\text{LCD}}$ is designed as follows, 
\begin{equation}
    {\cal L}_{\text{LCD}}({\bf \Theta}, {\bf \Theta}^{-}) = \mathbb{E} \left[d\left(\textbf{{\textit{f}}}_{{\bf \Theta}}(\mathbf{z}_{n+k}, t_{n+k}, w, \mathbf{c}), \textbf{{\textit{f}}}_{{\bf \Theta}^{-}}(\hat{\mathbf{z}}_{n}, t_n, w, \mathbf{c})\right) \right],
\end{equation}
where $d(\cdot, \cdot)$ is a distance measuring function, such as L2 loss or Huber loss~\cite{huber}. As discussed in~\cref{subsection: Preliminaries}, the target network ${\bf \Theta}^{-}$ is updated with the exponential moving average (EMA) of the trainable parameters of the online network ${\bf \Theta}$. Here we define the teacher network ${\bf \Theta}^{*}$ as the pre-trained motion latent diffusion model, \textit{i.e.}, MLD~\cite{mld}. According to~\cite{lcm}, the online network and target network are initialized with the parameters of the teacher network. During the inference phase, as shown in~\cref{figures:qualitative_t2m_v2}, our MotionLCM can generate high-quality motions with one-step sampling and achieve the fastest runtime (\textbf{$\sim$30ms per motion sequence}) compared to other motion diffusion models~\cite{mdm, mld}.

\subsection{Controllable Motion Generation in Latent Space}
\label{subsection: Controllable Motion Generation in Latent Space}

After addressing the low-efficiency issue in the motion latent diffusion model~\cite{mld}, we delve into another exploration of real-time motion control. Inspired by the great success of ControlNet~\cite{controlnet} in controllable image generation~\cite{sd}, we introduce a motion ControlNet ${\bf \Theta}^{a}$ in the latent space of MotionLCM and initialize the motion ControlNet with a trainable copy of MotionLCM. Specifically, each layer in the motion ControlNet is appended with a zero-initialized linear layer for eliminating random noise in the initial training steps. To achieve an autoregressive motion generation paradigm, as depicted in~\cref{figures/teaser}, we define the motion control task as generating motions given the initial $\tau$ poses and textual description. As shown in ~\cref{figures/pipeline_v2}(b), the initial $\tau$ poses are defined by the trajectories of $K$ control joints, $\mathbf{g}^{1:\tau} = \{\mathbf{g}^i\}_{i=1}^{\tau}$, where $\mathbf{g}^i \in \mathbb{R}^{K \times 3}$ denotes the global absolute locations of each control joint. In our motion control pipeline, we design a Trajectory Encoder ${\bf \Theta}^{b}$ consisting of stacked transformer~\cite{attention} layers to encode the trajectory signals. We append a global token (\textit{i.e.}, [CLS]) before the start of the trajectory sequence as the output feature of the encoder, which is added to the noisy $\mathbf{z}_n$ and fed into the trainable motion ControlNet ${\bf \Theta}^{a}$. Under the guidance of motion ControlNet, MotionLCM predicts the denoised $\hat{\mathbf{z}}_0$ through the consistency function $\textbf{\textit{f}}_{{\bf \Theta}^{s}}$, where ${\bf \Theta}^{s}$ is the combination of ${\bf \Theta}^{a}$, ${\bf \Theta}^{b}$ and ${\bf \Theta}$. The following reconstruction loss ${\cal L}_\text{recon}$ optimizes the motion ControlNet ${\bf \Theta}^{a}$ and Trajectory Encoder ${\bf \Theta}^{b}$,
\begin{equation}
    {\cal L}_\text{recon}(\mathbf{\Theta}^{a}, \mathbf{\Theta}^{b})  = \mathbb{E} \left[ d \left( \textbf{{\textit{f}}}_{{\bf \Theta}^{s}} \left( \mathbf{z}_{n}, t_{n}, w, \mathbf{c}^{*} \right), \mathbf{z}_{0} \right) \right],
\end{equation}
where $\mathbf{c}^{*}$ includes the text condition and control guidance from the Trajectory Encoder and the motion ControlNet. However, during training, the sole reconstruction supervision in the latent space is insufficient. We argue this is because the controllable motion generation requires more detailed constraints, which cannot be effectively provided solely by the reconstruction loss in the latent space. Unlike previous methods like OmniControl~\cite{omnicontrol}, which directly diffuse in the motion space, allowing explicit supervision of control signals, effectively supervising control signals in the latent space is non-trivial. Therefore, we utilize the frozen VAE~\cite{vae} decoder ${\cal D}$ to decode the latent $\hat{\mathbf{z}}_{0}$ into the motion space, obtaining the predicted motion $\hat{\mathbf{x}}_0$, thereby introducing the control loss ${\cal L}_\text{control}$ as follows,
\begin{equation}
    {\cal L}_\text{control}(\mathbf{\Theta}^{a}, \mathbf{\Theta}^{b}) = \mathbb{E} \left[ \frac{\sum_{i} \sum_{j} m_{ij} || R(\hat{\mathbf{x}}_{0})_{ij} - R(\mathbf{x}_{0})_{ij}||_{2}^{2}}{\sum_{i} \sum_{j} m_{ij}} \right],
    \label{eq:control_loss}
\end{equation}
where $R(\cdot)$ converts the joint local positions to global absolute locations and $m_{ij} \in \{0, 1\}$ is the binary joint mask at frame $i$ for the joint $j$. Then we optimize the motion ControlNet $\mathbf{\Theta}^{a}$ and Trajectory Encoder $\mathbf{\Theta}^{b}$ with the overall objective,
\begin{equation}
    \mathbf{\Theta}^{a}, \mathbf{\Theta}^{b} = \underset{\mathbf{\Theta}^{a}, \mathbf{\Theta}^{b}}{\arg\min} ({\cal L_\text{recon}} + \lambda {\cal L_\text{{control}}}),
\end{equation}
where $\lambda$ is the weight to balance the two losses. This design enables explicit control signals in the vanilla motion space to further provide supervision for the generation process. Extensive experiments demonstrate that the introduced supervision is beneficial in improving motion control performance, which will be introduced in the following section.

%% file: sections/4_experiments.tex
\section{Experiments}

In this section, we first present the experimental setup details in \cref{subsection: Experimental setup}. Subsequently, we provide quantitative and qualitative comparisons to evaluate the effectiveness of our proposed MotionLCM framework in \cref{subsection: Comparisons on Text-to-motion} and \cref{subsection: Comparisons on Controllable Motion Generation}. Finally, we conduct comprehensive ablation studies on MotionLCM in \cref{subsection: Ablation Studies}.

\subsection{Experimental setup}
\label{subsection: Experimental setup}

\noindent \textbf{Datasets.} We experiment on the popular HumanML3D~\cite{humanml3d} dataset, featuring 14,616 unique human motion sequences with 44,970 textual descriptions. For a fair comparison with previous methods~\cite{humanml3d, mdm, mld, motiondiffuse, temos}, we take the redundant motion representation, including root velocity, root height, local joint positions, velocities, rotations in root space, and the foot contact binary labels.

\noindent \textbf{Evaluation metrics.} We extend the evaluation metrics of previous works~\cite{humanml3d, omnicontrol, mld}. \textbf{(1) Time cost:} We follow~\cite{mld} to report the \underline{A}verage \underline{I}nference \underline{T}ime per \underline{S}entence (AITS) to evaluate the inference efficiency of models. \textbf{(2) Motion quality:} Frechet Inception Distance (FID) is adopted as a principal metric to evaluate the feature distributions between the generated and real motions. The feature extractor employed is from~\cite{humanml3d}. \textbf{(3) Motion diversity:} MultiModality (MModality) measures the generation diversity conditioned on the same text and Diversity calculates variance through features~\cite{humanml3d}. \textbf{(4) Condition matching:} Following~\cite{humanml3d}, we calculate the motion-retrieval precision (R-Precision) to report the text-motion Top-1/2/3 matching accuracy and Multimodal Distance (MM Dist) calculates the mean distance between motions and texts. \textbf{(5) Control error:} Trajectory error (Traj. err.) quantifies the ratio of unsuccessful trajectories, characterized by any control joint location error surpassing a predetermined threshold. Location error (Loc. err.) represents the unsuccessful joints. Average error (Avg. err.) denotes the mean location error of the control joints.

\noindent \textbf{Implementation details.} Our baseline motion diffusion model is based on MLD~\cite{mld}. We reproduce MLD with higher performance. Unless otherwise specified, all our experiments are conducted on this model. For MotionLCM, we employ the AdamW~\cite{adamw} optimizer for 96K iterations using a cosine decay learning rate scheduler and 1K iterations of linear warm-up. A batch size of 256 and a learning rate of 2e-4 are used. We set the training guidance scale range as $[w_\text{min}, w_\text{max}] = [5, 15]$, with the testing guidance scale set to 7.5, and adopt the EMA rate $\mu = 0.95$ by default. We use the DDIM~\cite{ddim} solver with skipping interval $k = 20$ and choose the Huber~\cite{huber} loss as the distance measuring function $d$. For motion ControlNet, we use the AdamW~\cite{adamw} optimizer for 192K iterations with 1K iterations of linear warm-up. The batch size and learning rate are set to 128 and 1e-4. The learning rate scheduler is the same as the first stage. For the training objective, we employ $d$ as the L2 loss and set the control loss weight $\lambda$ to 1.0 by default. We set the control ratio $\tau$ as 0.25 and the number of control joints as $K=6$ (\textit{i.e.}, \textit{Pelvis}, \textit{Left foot}, \textit{Right foot}, \textit{Head}, \textit{Left wrist}, and \textit{Right wrist}) in both training and testing. We implement our model using PyTorch~\cite{torch} with training on an NVIDIA RTX 4090 GPU and testing on a Tesla V100 GPU.

\subsection{Comparisons on Text-to-motion}
\label{subsection: Comparisons on Text-to-motion}

\input{tables/results_of_humanml3d_t2m}

\begin{figure}[t]
\begin{center}
\includegraphics[width=\linewidth]{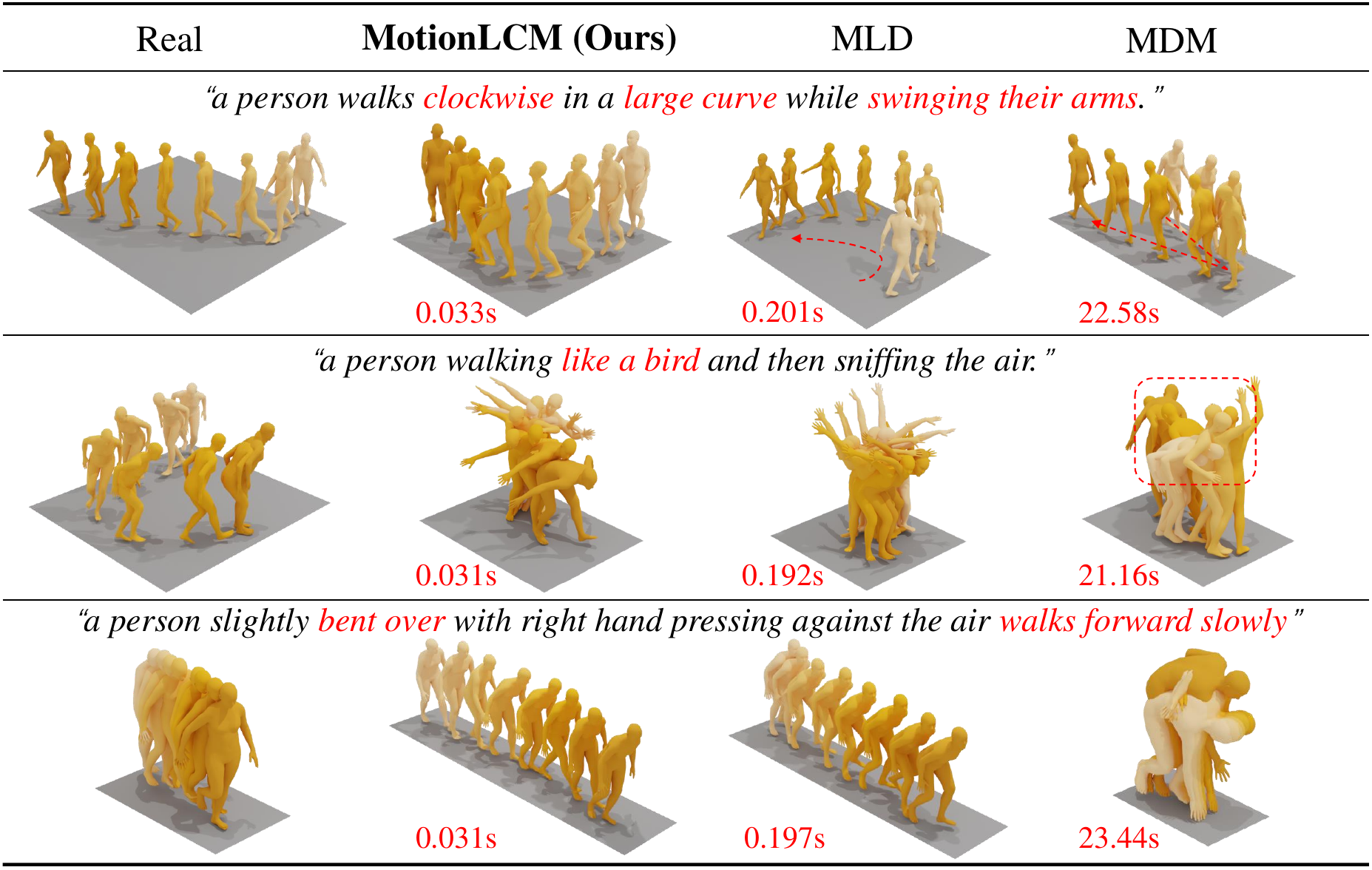}
\end{center}
\caption{Qualitative comparison of the state-of-the-art methods in the text-to-motion task. With only one-step inference, MotionLCM achieves the fastest motion generation while producing high-quality movements that closely match the textual descriptions.}
\label{figures:qualitative_t2m_v2}
\end{figure}

\noindent In the following part, we first evaluate our MotionLCM on the text-to-motion (T2M) task. We compare our method with some T2M baselines on HumanML3D \cite{humanml3d} with suggested metrics~\cite{humanml3d} under the 95\% confidence interval from 20 times running. As MotionLCM is based on MLD, we mainly focus on the performance compared with MLD. For evaluating time efficiency, we compare the Average Inference Time per Sentence (AITS) with TEMOS~\cite{temos}, T2M~\cite{humanml3d}, MDM~\cite{mdm}, MotionDiffuse~\cite{motiondiffuse} and MLD~\cite{mld}. The results are borrowed from MLD~\cite{mld}. The deterministic methods~\cite{seq2seq, jl2p, hier, t2g}, are unable to produce diverse results from a single input text and thus we leave their MModality metrics empty. For the quantitative results, as shown in \cref{table:results_of_humanml3d_t2m}, our MotionLCM boasts an impressive real-time runtime efficiency, averaging around \textbf{30ms per motion sequence} during inference. This performance exceeds that of previous diffusion-based methods~\cite{mdm, motiondiffuse, mld} and even surpasses MLD~\cite{mld} by an order of magnitude. Furthermore, despite employing only one-step inference, our MotionLCM can approximate or even surpass the performance of MLD~\cite{mld} (DDIM~\cite{ddim} 50 steps). With two-step inference, we achieve the best R-Precision and MM Dist metrics, while increasing the sampling steps to four yields the best FID. The above results demonstrate the effectiveness of latent consistency distillation. For the qualitative results, as shown in~\cref{figures:qualitative_t2m_v2}, MotionLCM not only accelerates motion generation to real-time speed but also delivers high-quality outputs, closely aligning with the textual descriptions.

\subsection{Comparisons on Controllable Motion Generation}
\label{subsection: Comparisons on Controllable Motion Generation}

\input{tables/results_of_humanml3d_t_ctrl}

\begin{figure}[t]
\begin{center}
\includegraphics[width=\linewidth]{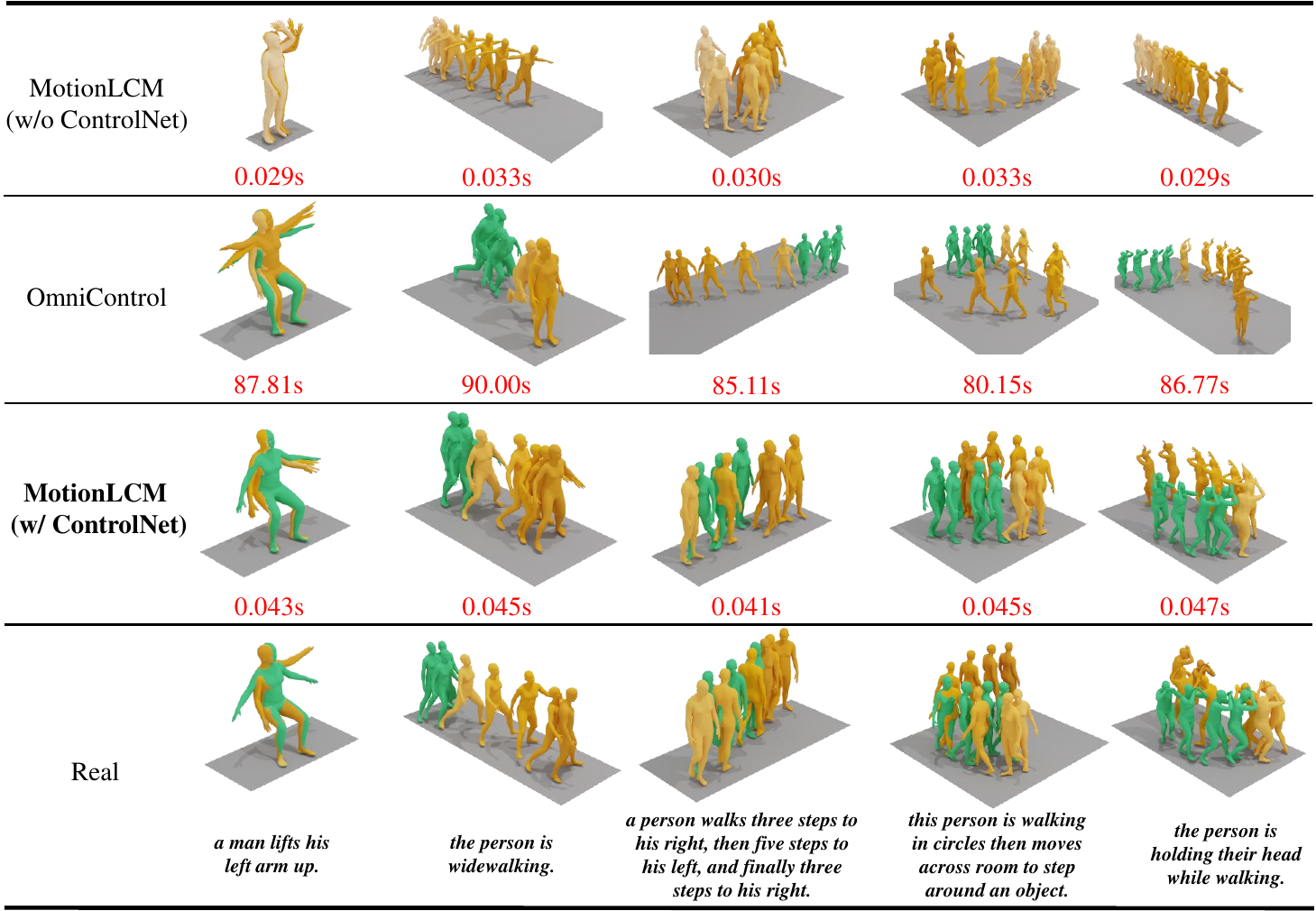}
\end{center}
\caption{Qualitative comparison of the state-of-the-art methods in the motion control task. We provide the visualized motion results and real references from five prompts. Compared to OmniControl~\cite{omnicontrol}, MotionLCM with ControlNet not only generates the initial poses that accurately follow the given multi-joint trajectories (\ie, the green poses in real references) but also produces motions that closely align with the texts.}
\label{figures:qualitative_tc_v2}
\end{figure}

As shown in~\cref{table:results_of_humanml3d_t_ctrl}, we compare our MotionLCM with OmniControl~\cite{omnicontrol} and MLD~\cite{mld}. We observe that OmniControl struggles with multi-joint control and falls short in both generation quality and control performance compared to MotionLCM. To verify the effectiveness of the latent generated by our MotionLCM for training motion ControlNet, we conducted the following two sets of experiments: ``LC'' and ``MC'', which indicate introducing control supervision in the latent space and motion space. It can be observed that under the same experimental settings, MotionLCM maintains higher fidelity and significantly outperforms MLD~\cite{mld} in motion control performance. This demonstrates that the latent generated by MotionLCM is more effective for training motion ControlNet compared to MLD~\cite{mld}. In terms of inference speed, MotionLCM (1-step) is \textbf{1929}$\times$ faster compared to OmniControl~\cite{omnicontrol} and \textbf{13}$\times$ faster than MLD~\cite{mld}. For qualitative results, as shown in~\cref{figures:qualitative_tc_v2}, OmniControl fails to control the initial poses in the second example and does not generate motion that aligns with the text in the third case. However, our MotionLCM not only adheres to the control of the initial poses but also generates motions that match the textual descriptions.

\subsection{Ablation Studies}
\label{subsection: Ablation Studies}

\input{tables/results_of_ablation_lcm}

\noindent \textbf{Impact of the hyperparameters of training MotionLCM.} We conduct a comprehensive analysis of the training hyperparameters of MotionLCM, including the training guidance scale range $[w_{\text{min}}, w_{\text{max}}]$, EMA rate $\mu$, skipping interval $k$, and the type of loss. We summarize the evaluation results based on one-step inference in~\cref{table:results_of_ablation_lcm}. We find out that using a dynamic training guidance scale (\textit{e.g.}, $w \in [5, 15]$) during training leads to an improvement in model performance compared to using a static training guidance scale (\textit{e.g.}, $w=7.5$). Additionally, an excessively large range for the training guidance scale can also negatively impact the performance of the model (\textit{e.g.}, $w \in [2, 18]$). Regarding the EMA rate $\mu$, we observe that the larger the value of $\mu$, the better the performance of the model. This indicates that maintaining a slower update rate for the target network ${\bf \Theta}^{-}$ helps improve the performance of latent consistency distillation. When the skipping interval $k$ continues to increase, the performance of the distillation model improves progressively, but larger values of $k$ (\textit{e.g.}, $k=50$) may result in inferior results. As for the type of loss, the Huber loss~\cite{huber} significantly outperforms the L2 loss, demonstrating its superior robustness.

\noindent \textbf{Impact of control loss weights $\lambda$.} To verify the impact of different control loss weights $\lambda$ on the control performance of MotionLCM, we report the experimental results in~\cref{table:results_of_ablation_t_ctrl_weight}. We also include experiments of MotionLCM without ControlNet (\textit{i.e.}, only text-to-motion) for comparison. We found a significant improvement in control-related metrics (\textit{e.g.}, Loc. err.) after introducing motion ControlNet (\textit{i.e.}, $\lambda = 0$). Furthermore, control performance can be further improved by introducing control loss (\textit{i.e.}, $\lambda > 0$). Increasing the weight $\lambda$ enhances control performance but leads to a decline in the generation quality, which is reflected in higher FID scores. To balance these two aspects, we adopt $\lambda=1$ as our default setting for training motion ControlNet.

\noindent \textbf{Impact of different control ratios $\tau$ and number of control joints $K$.} In~\cref{table:results_of_ablation_t_ctrl_tau_and_k}, we present the results of all models with the testing control ratio as 0.25 and keep the number of control joints $K$ equal in both training and testing. We found that the model with the fixed training control ratio (\textit{i.e.}, $\tau=0.25$) performs better compared to a dynamic ratio (\textit{e.g.}, $\tau \in [0.1, 0.5]$), and we discover that our model maintains good performance when incorporating additional redundant control signals, such as whole-body joints with $K=22$. 

\input{tables/results_of_ablation_t_ctrl_weight}

\input{tables/results_of_ablation_t_ctrl_tau_and_k}

%% file: tables/results_of_humanml3d_t2m.tex
\begin{table}[t]

\centering

\caption{Comparison of text-conditional motion synthesis on HumanML3D~\cite{humanml3d} dataset. We compute suggested metrics following~\cite{humanml3d}. We repeat the evaluation 20 times for each metric and report the average with a 95\% confidence interval. ``$\rightarrow$'' indicates that the closer to the real data, the better. \textbf{Bold} and \underline{underline} indicate the best and the second best result. ``$^{*}$'' denotes the reproduced version of MLD~\cite{mld}. The MotionLCM in \textbf{one-step inference (30ms)} surpasses all state-of-the-art models.}

\resizebox{\textwidth}{!}{

\scriptsize

\setlength\tabcolsep{3pt}

\begin{tabular}{lcccccccc}

\toprule

\multirow{2}{*}{Methods} & \multirow{2}{*}{AITS $\downarrow$} & \multicolumn{3}{c}{R-Precision $\uparrow$} & \multirow{2}{*}{FID $\downarrow$} & \multirow{2}{*}{MM Dist $\downarrow$} & \multirow{2}{*}{Diversity $\rightarrow$} & \multirow{2}{*}{MModality $\uparrow$}  \\

\cmidrule{3-5} & & Top 1 & Top 2 & Top 3 &  \\  \midrule

Real & - & $0.511^{\pm .003}$ & $0.703^{\pm .003}$ & $0.797^{\pm .002}$ & $0.002^{\pm .000}$ & $2.974^{\pm .008}$ & $9.503^{\pm .065}$ & - \\ \midrule

Seq2Seq~\cite{seq2seq} & - & $0.180^{\pm .002}$ & $0.300^{\pm .002}$ & $0.396^{\pm .002}$ & $11.75^{\pm .035}$ & $5.529^{\pm .007}$ & $6.223^{\pm .061}$ & - \\
JL2P~\cite{jl2p} & - & $0.246^{\pm .002}$ & $0.387^{\pm .002}$ & $0.486^{\pm .002}$ & $11.02^{\pm .046}$ & $5.296^{\pm .008}$ & $7.676^{\pm .058}$ & - \\
T2G~\cite{t2g} & - &$0.165^{\pm .001}$ & $0.267^{\pm .002}$ & $0.345^{\pm .002}$ & $7.664^{\pm .030}$ & $6.030^{\pm .008}$ & $6.409^{\pm .071}$ & - \\
Hier~\cite{hier} & - & $0.301^{\pm .002}$ & $0.425^{\pm .002}$ & $0.552^{\pm .004}$ & $6.532^{\pm .024}$ & $5.012^{\pm .018}$ & $8.332^{\pm .042}$ & - \\
TEMOS~\cite{temos} & 0.017 & $0.424^{\pm .002}$ & $0.612^{\pm .002}$ & $0.722^{\pm .002}$ & $3.734^{\pm .028}$ & $3.703^{\pm .008}$ & $8.973^{\pm .071}$ & $0.368^{\pm .018}$ \\
T2M~\cite{humanml3d} & 0.038 & $0.457^{\pm .002}$ & $0.639^{\pm .003}$ & $0.740^{\pm .003}$ & $1.067^{\pm .002}$ & $3.340^{\pm .008}$ & $9.188^{\pm .002}$ & $2.090^{\pm .083}$ \\
MDM~\cite{mdm} & 24.74 & $0.320^{\pm .005}$ & $0.498^{\pm .004}$ & $0.611^{\pm .007}$ & $0.544^{\pm .044}$ & $5.566^{\pm .027}$ & $\textbf{9.559}^{\pm .086}$ & $\textbf{2.799}^{\pm .072}$ \\
MotionDiffuse~\cite{motiondiffuse} & 14.74 & $0.491^{\pm .001}$ & $0.681^{\pm .001}$ & $0.782^{\pm .001}$ & $0.630^{\pm .001}$ & $3.113^{\pm .001}$ & $\underline{9.410}^{\pm .049}$ & $1.553^{\pm .042}$ \\
MLD~\cite{mld} & 0.217 & $0.481^{\pm .003}$ & $0.673^{\pm .003}$ & $0.772^{\pm .002}$ & $0.473^{\pm .013}$ & $3.196^{\pm .010}$ & $9.724^{\pm .082}$ & $\underline{2.413}^{\pm .079}$ \\ \midrule

\textbf{MLD}$^*$~\cite{mld} & 0.225 & $\underline{0.504}^{\pm .002}$ & $0.698^{\pm .003}$ & $0.796^{\pm .002}$ & $0.450^{\pm .011}$ & $3.052^{\pm .009}$ & $9.634^{\pm .064}$ & $2.267^{\pm .082}$ \\ \midrule

\textbf{MotionLCM (1-step)} & \textbf{0.030} & $0.502^{\pm .003}$ & $\underline{0.701}^{\pm .002}$ & $\underline{0.803}^{\pm .002}$ & $0.467^{\pm .012}$ & $3.022^{\pm .009}$ & $9.631^{\pm .066}$ & $2.172^{\pm .082}$ \\
\textbf{MotionLCM (2-step)} & \underline{0.035} & $\textbf{0.505}^{\pm .003}$ & $\textbf{0.705}^{\pm .002}$ & $\textbf{0.805}^{\pm .002}$ & $\underline{0.368}^{\pm .011}$ & $\textbf{2.986}^{\pm .008}$ & $9.640^{\pm .052}$ & $2.187^{\pm .094}$ \\
\textbf{MotionLCM (4-step)} & 0.043 & $0.502^{\pm .003}$ & $0.698^{\pm .002}$ & $0.798^{\pm .002}$ & $\textbf{0.304}^{\pm .012}$ & $\underline{3.012}^{\pm .007}$ & $9.607^{\pm .066}$ & $2.259^{\pm .092}$ \\ \bottomrule

\end{tabular}

}

\label{table:results_of_humanml3d_t2m}

\end{table}

%% file: tables/results_of_humanml3d_t_ctrl.tex
\begin{table}[t]

\centering

\caption{Comparison of motion control on HumanML3D~\cite{humanml3d} dataset. \textbf{Bold} indicates the best result. Our MotionLCM outperforms OmniControl~\cite{omnicontrol} and MLD~\cite{mld} regarding generation quality, control performance, and inference speed. ``LC'' and ``MC'' refer to the control supervision introduced in the latent space and motion space.}

\resizebox{\textwidth}{!}{

\scriptsize

\setlength\tabcolsep{3pt}

\begin{tabular}{lccccccc}

\toprule

\multirow{2}{*}{Methods} & AITS $\downarrow$ & FID $\downarrow$ & R-Precision $\uparrow$ & Diversity $\rightarrow$ & Traj. err. $\downarrow$ & Loc. err. $\downarrow$ & Avg. err. $\downarrow$ \\

& & & Top 3 & & (50cm) & (50cm) & \\ \midrule

Real & - & 0.002 & 0.797 & 9.503 & 0.0000 & 0.0000 & 0.0000 \\ \midrule

OmniControl~\cite{omnicontrol} & 81.00 & 2.328 & 0.557 & 8.867 & 0.3362 & 0.0322 & 0.0977 \\ \midrule

MLD~\cite{mld} (LC) & 0.552 & 0.469 & 0.723 & \textbf{9.476} & 0.4230 & 0.0653 & 0.1690 \\ 
MotionLCM (1-step, LC) & \textbf{0.042} & 0.319 & 0.752 & 9.424 & 0.2986 & 0.0344 & 0.1410 \\
MotionLCM (2-step, LC) & 0.047 & \textbf{0.315} & \textbf{0.770} & 9.427 & 0.2840 & 0.0328 & 0.1365 \\
MotionLCM (4-step, LC) & 0.063 & 0.328 & 0.745 & 9.441 & 0.2973 & 0.0339 & 0.1398 \\ \midrule

MLD~\cite{mld} (LC\&MC) & 0.552 & 0.555 & 0.754 & 9.373 & 0.2722 & 0.0215 & 0.1265 \\
MotionLCM (1-step, LC\&MC) & \textbf{0.042} & 0.419 & 0.756 & 9.390 & 0.1988 & 0.0147 & 0.1127 \\
MotionLCM (2-step, LC\&MC) & 0.047 & 0.397 & 0.759 & 9.469 & \textbf{0.1960} & \textbf{0.0143} & \textbf{0.1092} \\
MotionLCM (4-step, LC\&MC) & 0.063 & 0.444 & 0.753 & 9.355 & 0.2089 & 0.0172 & 0.1140 \\ \bottomrule

\end{tabular}

}

\label{table:results_of_humanml3d_t_ctrl}

\end{table}

%% file: tables/results_of_ablation_lcm.tex
\begin{table}[t]

\centering

\caption{Ablation study on different training guidance scale ranges $[w_{\text{min}}, w_{\text{max}}]$, EMA rates $\mu$, skipping intervals $k$ and types of loss. We use metrics in \cref{table:results_of_humanml3d_t2m} and adopt a one-step inference setting with the testing CFG scale of 7.5 for fair comparison.}

\resizebox{0.9\textwidth}{!}{

\scriptsize

\setlength\tabcolsep{3pt}

\begin{tabular}{lccccc}

\toprule

\multirow{2}{*}{Methods} & R-Precision $\uparrow$ & \multirow{2}{*}{FID $\downarrow$} & \multirow{2}{*}{MM Dist $\downarrow$} & \multirow{2}{*}{Diversity $\rightarrow$} & \multirow{2}{*}{MModality $\uparrow$}  \\ 

& Top 1 & & & &  \\ \midrule

Real & $0.511^{\pm .003}$ & $0.002^{\pm .000}$ & $2.974^{\pm .008}$ & $9.503^{\pm .065}$ & - \\ \midrule

MotionLCM ($w \in [5, 15]$) & $\textbf{0.502}^{\pm .003}$ & $0.467^{\pm .012}$ & $3.022^{\pm .009}$ & $9.631^{\pm .066}$ & $2.172^{\pm .082}$ \\
MotionLCM ($w \in [2, 18]$) & $0.497^{\pm .003}$ & $0.481^{\pm .009}$ & $3.030^{\pm .010}$ & $9.644^{\pm .073}$ & $2.226^{\pm .091}$ \\
MotionLCM ($w = 7.5$) & $0.486^{\pm .002}$ & $0.479^{\pm .009}$ & $3.094^{\pm .009}$ & $9.610^{\pm .072}$ & $2.320^{\pm .097}$ \\ \midrule

MotionLCM ($\mu = 0.95$) & $\textbf{0.502}^{\pm .003}$ & $0.467^{\pm .012}$ & $3.022^{\pm .009}$ & $9.631^{\pm .066}$ & $2.172^{\pm .082}$ \\
MotionLCM ($\mu = 0.50$) & $0.498^{\pm .003}$ & $0.478^{\pm .009}$ & $3.022^{\pm .010}$ & $9.655^{\pm .071}$ & $2.188^{\pm .087}$ \\
MotionLCM ($\mu = 0$) & $0.499^{\pm .003}$ & $0.505^{\pm .008}$ & $3.018^{\pm .009}$ & $9.706^{\pm .070}$ & $2.123^{\pm .085}$ \\ \midrule

MotionLCM ($k = 50$) & $0.488^{\pm .003}$ & $0.547^{\pm .011}$ & $3.096^{\pm .010}$ & $\textbf{9.511}^{\pm .074}$ & $\textbf{2.324}^{\pm .091}$ \\
MotionLCM ($k = 20$) & $\textbf{0.502}^{\pm .003}$ & $0.467^{\pm .012}$ & $3.022^{\pm .009}$ & $9.631^{\pm .066}$ & $2.172^{\pm .082}$ \\
MotionLCM ($k = 10$) & $0.497^{\pm .003}$ & $0.449^{\pm .009}$ & $\textbf{3.017}^{\pm .010}$ & $9.693^{\pm .075}$ & $2.133^{\pm .086}$ \\ 
MotionLCM ($k = 5$) & $0.488^{\pm .003}$ & $\textbf{0.438}^{\pm .009}$ & $3.044^{\pm .009}$ & $9.647^{\pm .074}$ & $2.147^{\pm .083}$ \\ 
MotionLCM ($k = 1$) & $0.442^{\pm .002}$ & $0.635^{\pm .011}$ & $3.255^{\pm .008}$ & $9.384^{\pm .080}$ & $2.146^{\pm .075}$ \\ \midrule

MotionLCM (w/ Huber) & $\textbf{0.502}^{\pm .003}$ & $0.467^{\pm .012}$ & $3.022^{\pm .009}$ & $9.631^{\pm .066}$ & $2.172^{\pm .082}$ \\
MotionLCM (w/  L2) & $0.486^{\pm .002}$ & $0.622^{\pm .010}$ & $3.114^{\pm .009}$ & $9.573^{\pm .069}$ & $2.218^{\pm .086}$ \\ \bottomrule

\end{tabular}

}

\label{table:results_of_ablation_lcm}

\end{table}

%% file: tables/results_of_ablation_t_ctrl_weight.tex
\begin{table}[t]

\centering

\caption{Ablation study on different control loss weights $\lambda$. We present the results of (1, 2, 4)-step inference. We add the MotionLCM without ControlNet for comparison.}

\resizebox{0.92\textwidth}{!}{

\scriptsize

\setlength\tabcolsep{3pt}

\begin{tabular}{lcccccc}

\toprule

\multirow{2}{*}{Methods} & FID $\downarrow$ & R-Precision $\uparrow$ & Diversity $\rightarrow$ & Traj. err. $\downarrow$ & Loc. err. $\downarrow$ & Avg. err. $\downarrow$ \\

& & Top 3 & & (50cm) & (50cm) & \\ \midrule

Real & 0.002 & 0.797 & 9.503 & 0.0000 & 0.0000 & 0.0000 \\ \midrule

MotionLCM (1-step, w/o control) & 0.467 & 0.803 & 9.631 & 0.7605 & 0.2302 & 0.3493 \\
MotionLCM (2-step, w/o control) & 0.368 & \textbf{0.805} & 9.640 & 0.7646 & 0.2214 & 0.3386 \\
MotionLCM (4-step, w/o control) & \textbf{0.304} & 0.798 & 9.607 & 0.7739 & 0.2207 & 0.3359 \\ \midrule

MotionLCM (1-step, $\lambda=0$) & 0.319 & 0.752 & 9.424 & 0.2986 & 0.0344 & 0.1410 \\
MotionLCM (2-step, $\lambda=0$) & 0.315 & 0.770 & 9.427 & 0.2840 & 0.0328 & 0.1365 \\
MotionLCM (4-step, $\lambda=0$) & 0.328 & 0.745 & 9.441 & 0.2973 & 0.0339 & 0.1398 \\ \midrule

MotionLCM (1-step, $\lambda=0.1$) & 0.344 & 0.753 & 9.386 & 0.2711 & 0.0275 & 0.1310 \\
MotionLCM (2-step, $\lambda=0.1$) & 0.324 & 0.759 & 9.428 & 0.2631 & 0.0256 & 0.1268 \\
MotionLCM (4-step, $\lambda=0.1$) & 0.357 & 0.743 & 9.415 & 0.2713 & 0.0268 & 0.1309 \\ \midrule

MotionLCM (1-step, $\lambda=1.0$) & 0.419 & 0.756 & 9.390 & 0.1988 & 0.0147 & 0.1127 \\
MotionLCM (2-step, $\lambda=1.0$) & 0.397 & 0.759 & 9.469 & 0.1960 & 0.0143 & 0.1092 \\
MotionLCM (4-step, $\lambda=1.0$) & 0.444 & 0.753 & 9.355 & 0.2089 & 0.0172 & 0.1140 \\ \midrule

MotionLCM (1-step, $\lambda=10.0$) & 0.636 & 0.744 & 9.479 & \textbf{0.1465} & \textbf{0.0097} & \textbf{0.0967} \\
MotionLCM (2-step, $\lambda=10.0$) & 0.551 & 0.757 & 9.569 & 0.1590 & 0.0107 & 0.0987 \\
MotionLCM (4-step, $\lambda=10.0$) & 0.568 & 0.742 & \textbf{9.486} & 0.1723 & 0.0132 & 0.1045 \\ \bottomrule

\end{tabular}

}

\label{table:results_of_ablation_t_ctrl_weight}

\end{table}

%% file: tables/results_of_ablation_t_ctrl_tau_and_k.tex
\begin{table}[t]

\centering

\caption{Ablation study on different control ratios $\tau$ and number of control joints $K$. We report the results of (1, 2, 4)-step inference. ``$^{*}$'' is the default training setting.}

\resizebox{0.92\textwidth}{!}{

\scriptsize

\setlength\tabcolsep{3pt}

\begin{tabular}{lcccccc}

\toprule

\multirow{2}{*}{Methods} & FID $\downarrow$ & R-Precision $\uparrow$ & Diversity $\rightarrow$ & Traj. err. $\downarrow$ & Loc. err. $\downarrow$ & Avg. err. $\downarrow$ \\

& & Top 3 & & (50cm) & (50cm) & \\ \midrule

Real & 0.002 & 0.797 & 9.503 & 0.0000 & 0.0000 & 0.0000 \\ \midrule

MotionLCM$^{*}$ (1-step, $\tau=0.25$, $K=6$) & 0.419 & 0.756 & 9.390 & 0.1988 & 0.0147 & 0.1127 \\
MotionLCM$^{*}$ (2-step, $\tau=0.25$, $K=6$) & \textbf{0.397} & 0.759 & 9.469 & \textbf{0.1960} & 0.0143 & 0.1092 \\
MotionLCM$^{*}$ (4-step, $\tau=0.25$, $K=6$) & 0.444 & 0.753 & 9.355 & 0.2089 & 0.0172 & 0.1140 \\ \midrule

MotionLCM (1-step, $\tau \in [0.1, 0.25]$) & 0.456 & 0.757 & 9.477 & 0.2821 & 0.0234 & 0.1214 \\
MotionLCM (2-step, $\tau \in [0.1, 0.25]$) & 0.409 & \textbf{0.769} & 9.592 & 0.2707 & 0.0230 & 0.1179 \\
MotionLCM (4-step, $\tau \in [0.1, 0.25]$) & 0.457 & 0.757 & 9.540 & 0.2928 & 0.0256 & 0.1228 \\ \midrule

MotionLCM (1-step, $\tau \in [0.1, 0.5]$) & 0.448 & 0.763 & 9.538 & 0.2390 & 0.0182 & 0.1182 \\
MotionLCM (2-step, $\tau \in [0.1, 0.5]$) & 0.413 & 0.768 & 9.517 & 0.2349 & 0.0180 & 0.1153 \\
MotionLCM (4-step, $\tau \in [0.1, 0.5]$) & 0.446 & 0.753 & \textbf{9.498} & 0.2517 & 0.0199 & 0.1196  \\ \midrule

MotionLCM (1-step, $K=12$) & 0.412 & 0.753 & 9.412 & 0.2072 & 0.0110 & 0.1029 \\
MotionLCM (2-step, $K=12$) & 0.410 & 0.758 & 9.509 & 0.1979 & 0.0108 & 0.1000 \\
MotionLCM (4-step, $K=12$) & 0.442 & 0.755 & 9.380 & 0.2169 & 0.0132 & 0.1048 \\ \midrule

MotionLCM (1-step, $K=22 (\text{whole-body})$) & 0.436 & 0.748 & 9.379 & 0.2143 & 0.0083 & 0.0914 \\
MotionLCM (2-step, $K=22 (\text{whole-body})$) & 0.413 & 0.758 & 9.492 & 0.2061 & \textbf{0.0082} & \textbf{0.0881} \\
MotionLCM (4-step, $K=22 (\text{whole-body})$) & 0.461 & 0.745 & 9.459 & 0.2173 & 0.0097 & 0.0918 \\ \bottomrule

\end{tabular}

}

\label{table:results_of_ablation_t_ctrl_tau_and_k}

\end{table}

%% file: sections/5_conclusion.tex
\section{Conclusion}

This work proposes an efficient controllable motion generation framework, MotionLCM. By introducing latent consistency distillation, MotionLCM enjoys the trade-off between runtime efficiency and generation quality. Moreover, thanks to the motion ControlNet manipulation in the latent space, our method obtains excellent controlling ability with given conditions. Extensive experimental results show the effectiveness of the proposed method. As the VAE of MLD lacks explicit temporal modeling, the MotionLCM cannot achieve a good temporal explanation. Therefore, our future work will lie in developing a more explainable compression architecture for efficient motion control. 

\section*{Acknowledgements}

The research is supported by Shenzhen Ubiquitous Data Enabling Key Lab under grant ZDSYS20220527171406015 and CCF-Tencent Rhino-Bird Open Research Fund. This project is also supported by Shanghai Artificial Intelligence Laboratory. The author team would like to acknowledge Yiming Xie, Zhiyang Dou, and Shunlin Lu for their helpful technical discussions and suggestions.

%% file: sections/appendix.tex
\appendix
\maketitlesupplementary

In this supplementary material, we provide additional details and experiments not included in the main paper due to limitations in space.
\begin{itemize}
    \item \cref{section: Additional Experiments}: Additional experiments.
    \item \cref{section: More Qualitative Results}: Supplementary quantitative results.
    \item \cref{section: Metric Definitions}: Details of the evaluation metrics.
\end{itemize}

\section{Additional Experiments}
\label{section: Additional Experiments}

\subsection{Comparison to other ODE Solvers}

To validate the effectiveness of latent consistency distillation, we compare three ODE solvers (DDIM~\cite{ddim}, DPM~\cite{dpm1}, DPM$++$~\cite{dpm2}). The quantitative results shown in~\cref{table:results_of_ablation_solvers} demonstrate that our MotionLCM notably outperforms baseline methods. Moreover, unlike DDIM~\cite{ddim}, DPM~\cite{dpm1}, and DPM$++$~\cite{dpm2}, requiring more peak memory per sampling step when using CFG~\cite{cfg}, MotionLCM only requires one forward pass, saving both time and memory costs.

\input{tables/results_of_ablation_solvers}

\subsection{Impact of different testing CFGs}

\begin{wrapfigure}{r}{0.45\textwidth}
    \vspace{-2.0em}
    \includegraphics[width=\linewidth]{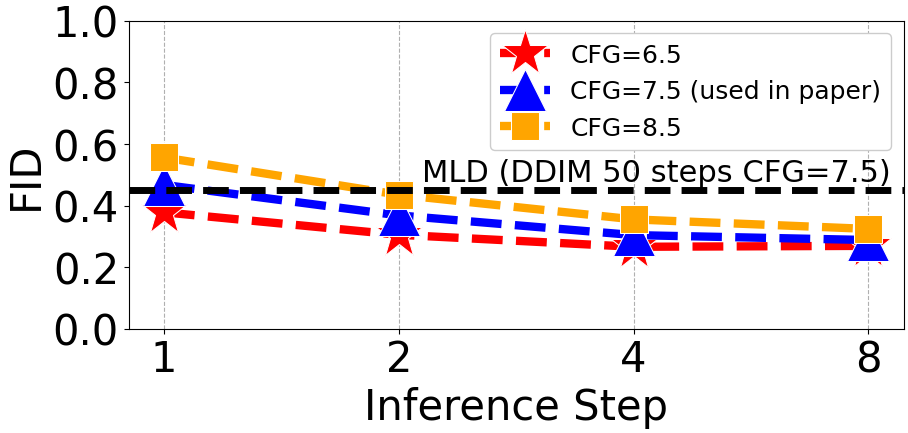}
    \caption{Comparison of testing CFGs}
    \label{figures/ablation_cfg_step}
    \vspace{-1.2em}
\end{wrapfigure}

As shown in~\cref{figures/ablation_cfg_step}, we provide an extensive ablation study on the testing CFG~\cite{cfg}. It can be observed that, under different testing CFGs, increasing the number of inference steps continuously improves the performance. However, further increasing the inference steps results in comparable performance but significantly increases the time cost.

\section{More Qualitative Results}
\label{section: More Qualitative Results}

\begin{figure}[!h]
    \centering
    \includegraphics[width=1.0\columnwidth]{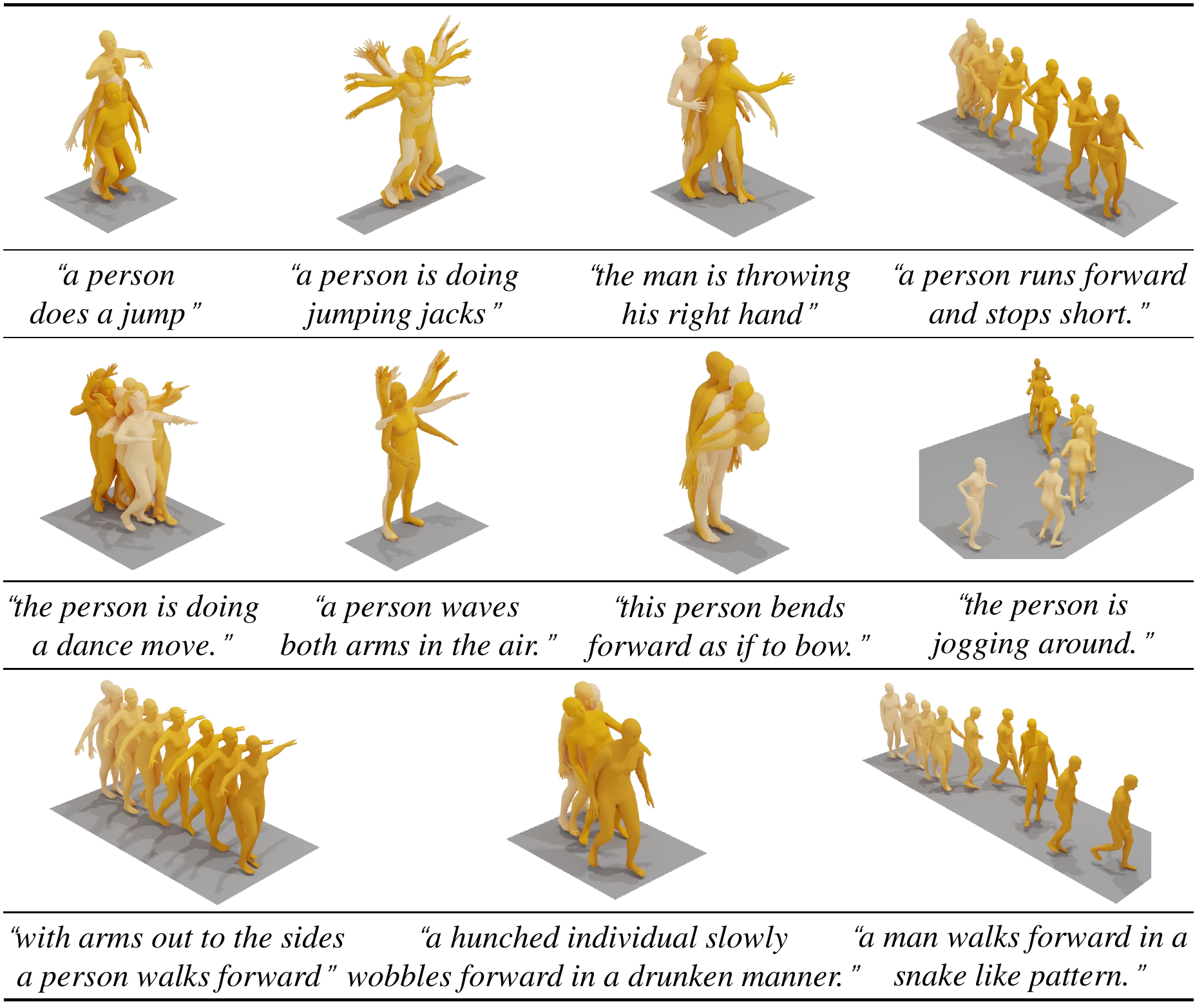}
    \caption{More qualitative results of MotionLCM on the text-to-motion task.}
    \label{figures/qualitative_t2m_supp}
\end{figure}

\begin{figure}[!h]
    \centering
    \includegraphics[width=1.0\columnwidth]{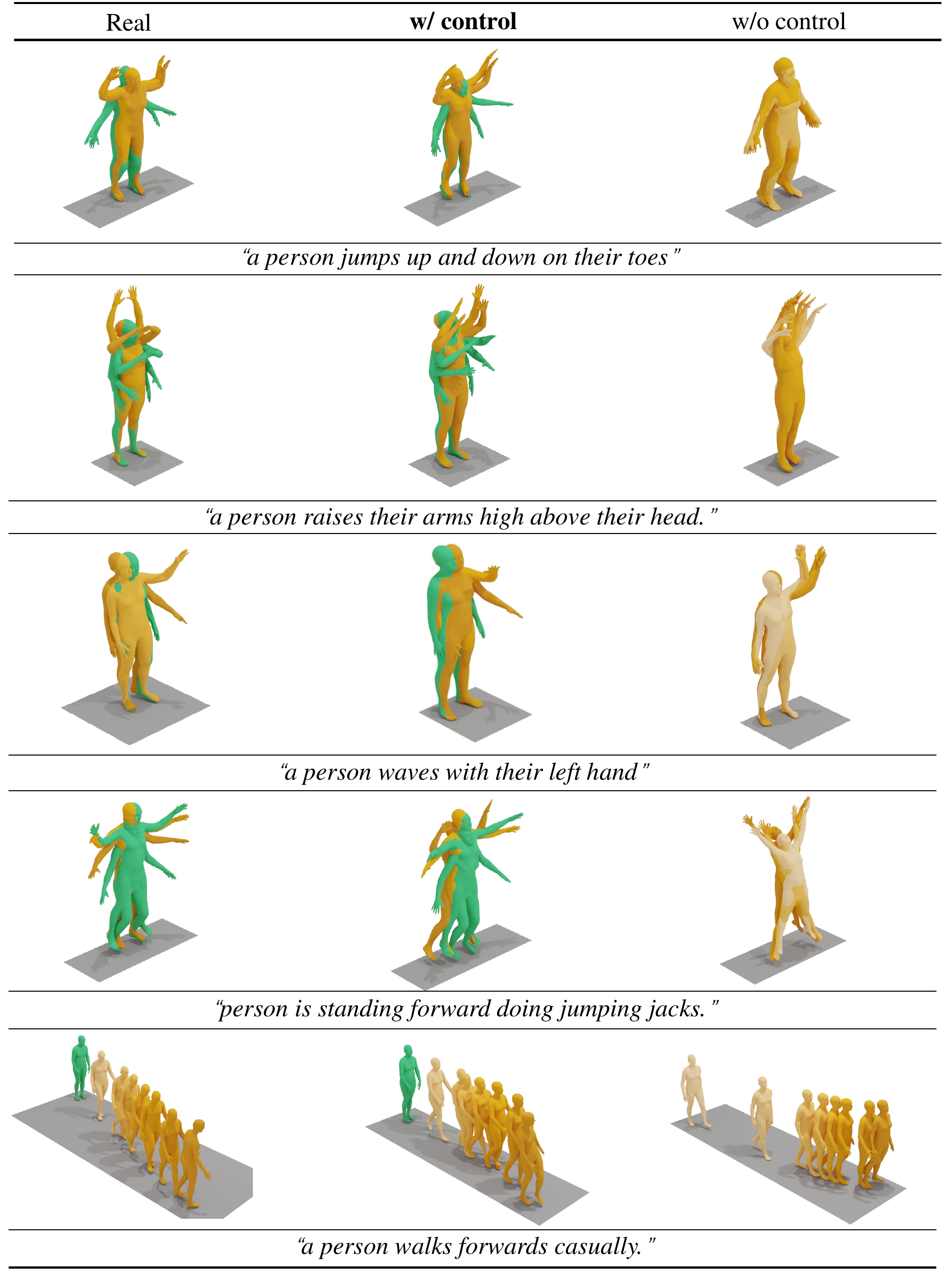}
    \caption{More qualitative results of MotionLCM on the motion control task.}
    \label{figures/qualitative_tc_supp}
\end{figure}

In this section, we provide more qualitative results of our MotionLCM. \cref{figures/qualitative_t2m_supp} presents more generation results on the text-to-motion task. \cref{figures/qualitative_tc_supp} displays additional visualization results on the motion control task. All videos shown in the figures can be found in the supplementary video (\textit{i.e.}, \textit{supp.mp4}).

\section{Metric Definitions}
\label{section: Metric Definitions}

\noindent \textbf{Time cost:} To assess the inference efficiency of models, we follow~\cite{mld} to report the Average Inference Time per Sentence (AITS) measured in seconds. We calculate AITS on the test set of HumanML3D~\cite{humanml3d} by setting the batch size to 1 and excluding the time cost for model and dataset loading parts.

\noindent \textbf{Motion quality:} Frechet Inception Distance (FID) measures the distributional difference between the generated and real motions, calculated using the feature extractor associated with a specific dataset, \textit{e.g.}, HumanML3D~\cite{humanml3d}.

\noindent \textbf{Motion diversity:} Following~\cite{action2motion, tm2t}, we report Diversity and MultiModality to evaluate the generated motion diversity. Diversity measures the variance of the generated motions across the whole set. Specifically, two subsets of the same size $S_d$ are randomly sampled from all generated motions with their extracted motion feature vectors $\{\mathbf{v}_1, ..., \mathbf{v}_{S_d}\}$ and $\{\mathbf{v}^{'}_1, ..., \mathbf{v}^{'}_{S_d}\}$. Diversity is defined as follows,
\begin{equation}
    \text{Diversity} = \frac{1}{S_d}\sum_{i=1}^{S_d}||\mathbf{v}_i - \mathbf{v}^{'}_i||_2.
\end{equation}
Different from Diversity, MultiModality (MModality) measures how much the generated motions diversify within each textual description. Specifically, a set of textual descriptions with size $C$ is randomly sampled from all descriptions. Then we randomly sample two subsets with the same size $I$ from all generated motions conditioned by the $c$-th textual description, with extracted feature vectors $\{\mathbf{v}_{c, 1}, ..., \mathbf{v}_{c, I}\}$ and $\{\mathbf{v}^{'}_{c, 1}, ..., \mathbf{v}^{'}_{c, I}\}$. MModality is formalized as follows,
\begin{equation}
    \text{MModality} = \frac{1}{C \times I}\sum_{c=1}^{C}\sum_{i=1}^{I}||\mathbf{v}_{c, i} - \mathbf{v}^{'}_{c, i}||_2.
\end{equation}

\noindent \textbf{Condition matching:} \cite{humanml3d} provides motion/text feature extractors to generate geometrically closed features for matched text-motion pairs and vice versa. Under this feature space, evaluating motion-retrieval precision (R-Precision) involves mixing the generated motion with 31 mismatched motions and then calculating the text-motion Top-1/2/3 matching accuracy. Multimodal Distance (MM Dist) calculates the mean distance between the generated motions and texts.

\noindent \textbf{Control error:} Following~\cite{omnicontrol}, we report Trajectory error, Location error, and Average error to assess the motion control performance. Trajectory error (Traj. err.) is defined as the proportion of unsuccessful trajectories, \textit{i.e.},  if a control joint in the generated motion exceeds a certain distance threshold from the corresponding joint in the given control trajectory, it is considered a failed trajectory. Similar to the Trajectory error, Location error (Loc. err.) is defined as the ratio of unsuccessful joints. In our experiments, we adopt 50cm as the distance threshold to calculate the Trajectory error and Location error. Average error (Avg. err.) denotes the mean distance between the control joint positions in the generated motion and those in the given control trajectory.

%% file: tables/results_of_ablation_solvers.tex
\begin{table}[h]

\vspace{-1.3em}

\centering

\caption{Quantitative results with the testing CFG scale $w=7.5$. MotionLCM notably outperforms baseline methods~\cite{ddim, dpm1, dpm2} on HumanML3D~\cite{humanml3d} dataset, demonstrating the effectiveness of latent consistency distillation. \textbf{Bold} indicates the best result.}

\resizebox{\textwidth}{!}{

\scriptsize

\renewcommand\arraystretch{1.55}

\setlength\tabcolsep{2.0pt}

\begin{tabular}{lcccccc}

\toprule

\multirow{2}{*}{Methods} & \multicolumn{3}{c}{R-Precision (Top 3) $\uparrow$} & \multicolumn{3}{c}{FID $\downarrow$} \\ 

\cmidrule(lr{0.2em}){2-4} \cmidrule(lr{0.2em}){5-7} & {1-Step} & 2-Step & {4-Step} & 1-Step & 2-Step & 4-Step \\ \hline

DDIM~\cite{ddim} & $0.651^{\pm .003}$ & $0.691^{\pm .002}$ & $0.765^{\pm .003}$ & $4.022^{\pm .043}$ & $2.802^{\pm .038}$ & $0.966^{\pm .018}$ \\

DPM~\cite{dpm1} & $0.651^{\pm .003}$ & $0.691^{\pm .002}$ & $0.777^{\pm .003}$ & $4.022^{\pm .043}$ & $2.798^{\pm .038}$ & $0.727^{\pm .015}$ \\

DPM$++$~\cite{dpm2} & $0.651^{\pm .003}$ & $0.691^{\pm .002}$ & $0.777^{\pm .003}$ & $4.022^{\pm .043}$ & $2.798^{\pm .038}$ & $0.684^{\pm .015}$ \\ \hline

\textbf{MotionLCM} & $\textbf{0.803}^{\pm .002}$ & $\textbf{0.805}^{\pm .002}$ & $\textbf{0.798}^{\pm .002}$ & $\textbf{0.467}^{\pm .012}$ & $\textbf{0.368}^{\pm .011}$ & $\textbf{0.304}^{\pm .012}$ \\ \bottomrule

\end{tabular}

}

\label{table:results_of_ablation_solvers}

\vspace{-2.0em}

\end{table}